\documentclass[conference]{IEEEtran}
\IEEEoverridecommandlockouts
\usepackage{cite}
\usepackage{amsmath,amssymb,amsfonts}
\usepackage{algorithmic}

\usepackage[linesnumbered,ruled]{algorithm2e}
\AtBeginEnvironment{algorithm}{}

\usepackage{graphicx}
\usepackage{textcomp}
\usepackage{xcolor}

\usepackage{tcolorbox}
\usepackage[frozencache,cachedir=.]{minted}

\definecolor{bg}{gray}{0.95}

\newcommand{\step}{$step$}

\newcommand{\console}[1]{\texttt{\small #1}}

\def\BibTeX{{\rm B\kern-.05em{\sc i\kern-.025em b}\kern-.08em
    T\kern-.1667em\lower.7ex\hbox{E}\kern-.125emX}}
\begin{document}

\title{Programming Manipulators by Instructions
}

\author{\IEEEauthorblockN{Rafael de la Guardia}
\IEEEauthorblockA{\textit{Intel Labs} \\
\textit{Intel Corporation Iberia, S.A.}\\
Madrid, Spain \\
rafael.de.la.guardia@intel.com}}

\maketitle

\begin{abstract}
We propose an instructions-based approach for robot programming where the programmer interacts with the robot by issuing simple commands in a scripting language, like python. Internally, these commands make use of pre-programmed motion and manipulation skills coordinated by a behaviour tree task controller. A knowledge graph keeps track of the state of the robot and the environment and of all the instructions given to the robot by the programmer. This allows to easily transform sequences of instructions into new skills that can be reused in the same or in other tasks. An advantage of this approach is that the programmer does not need to be located physically next to the robot, but can work remotely, either with a physical robot or with a digital twin. To demonstrate the concept, we show an interactive simulation of a robot manipulator in a pick and place scenario.
\end{abstract}


\section{Introduction}
Traditional robot programming relies on repeated expert demonstration, for example via kinesthetic teaching, or more recently, leveraging extended reality interfaces~\cite{ravichandar2020recent}. The data collected during demonstrations can be used to train models in various learning frameworks, which provide the robot not only with low‐level motion skills, but also high‐level symbolic reasoning skills obtained from the human programmer. However, programming robot skills takes significant expertise, and is expensive and difficult to do on the field. Moreover, performing complex tasks requires composition of multiple skills, possibly with different parameters and applied to different payloads. We argue that a minimum of base skills should be designed by experts, to allow end users of the robot to solve basic tasks in a domain of operation, and to enable them to gradually expand the capabilities of the robot by converting tasks into new skills. 
In the following section, we briefly describe related work in skills acquisition and skill-based control. Afterwards, we present an overview of our methodology, followed by an in-depth description of the innovative elements. We finalize by showing experimental results and conclusions.

\section{Related work}

\subsection{Skills acquisition}
~\cite{rovida_motion_2018} models skills as concurrent motion primitives that can be activated dynamically. ~\cite{celik2022specializing} model behavior with contextual skill libraries of motion primitives formalized by Mixtures of Experts. The context defines task properties, like reaching different goal positions or different friction parameters of an object to manipulate and the goal is to learn versatile skills to solve a given context. The review in ~\cite{tavassoli2023learning} focuses on Movement Primitives and Experience Abstraction, i.e., skills, as methods to reduce complex LfD problems in robotics to their minimum components, while ~\cite{si2021review} targets manipulation skill acquisition in the context of teleoperation. In the present work, we assume a relatively small library of skills is available. These skills are created by experts and designed to cover a large number of applications in a specific use case, such as pick and place. Our objective is to enable non-expert users to extend the library by composition of the existing skills in order to facilitate programming in their specific use case.

\subsection{Skills-based task control}
Following skills-based approach requires state machines or behaviour trees~\cite{mower2024ros} to coordinate the task execution. Behaviour trees (BTs) are control architectures which take the form of ordered directed trees~\cite{ogren2022behavior}. BTs are modular in the sense that every sub-tree, including the whole tree, can be reused meaningfully in another tree. This is a key property that we exploit in our framework. BTs can be reactive, meaning that whenever the current input is the same, they always select the same action regardless of input history. This lack of memory should not be confused with a lack of knowledge about the world, which is essential for robotic applications. In this work, we advocate for knowledge graphs as an internal world model that abstracts sensor input into a cohesive structure which can be queried by BTs. Doing so can be made consistent with reactiveness, so long as only the current state of this model is used to determine the action selected. A few recent works use similar skill-based approaches.
Skiros~\cite{mayr_skiros2_2023} features a layered hybrid control system, integrating explicit knowledge representation and task-level planning. It combines execution of skills using a behaviour tree and an explicit world model, reading ontologies and the vocabulary that is stored in scenes. Task-level plans are generated by combining goals in PDDL and problem descriptions based on the WM’s knowledge. ~\cite{styrud2023bebopcombiningreactive} generates BTs by building a reactive tree structure using a PDDL planner and then subsequently learns the BT parameters with Bayesian Optimization (BO). With the tree structure as a prior, BO can then focus on tuning parameters that are difficult to plan and reason about. ~\cite{mower2024ros} present a framework that enables non-experts to program robots using natural language prompts and contextual information retrieved from a ROS environment. New skills can be added to a library by first, demonstrating the skill via kinesthetic teaching or teleoperation, and then automatically generating a ROS service and a LLM-callable tool from a text description. ~\cite{jiang2023bridginglowlevelgeometryhighlevel} aligns geometric and semantic task specifications, by jointly using Vision-Language Models and Event Knowledge Graphs (EKGs) to process concepts and events in robot manipulation tasks. Semantic task specifications are obtained through EKGs to guide robot manipulation. A memory of successful manipulations is available that can be matched to current observation to retrieve semantic information and to specify the geometric constraints involved to execute the task.
~\cite{gustavsson2022combiningcontextawarenessplanning} learns task constraints and contexts to correctly design the BT using backchaining: starting from the goal, pre-conditions are iteratively expanded with actions that achieve them. Then, those actions’ unmet pre-conditions are expanded in the same way. 
At the same time, candidate post-conditions of an action are checked for compatibility against the pre-condition of the following action and discarded if conflicts are generated. In ~\cite{10000088}, 
a robot asks clarification questions to the user to resolve ambiguities and identify the correct targets to accurately complete the task. In the present work, we implement the Task Controller as a BT which is designed to select actions from a library of skills, loaded at the start of a programming session. In addition, new skills constructed by combining base skills can be selected using a new type of node that guides the controller to the appropriate actions to take following the execution of each base skill. Knowledge about the skills and the context in which they are used are stored in a Knowledge Graph.

\section{Method overview}

\begin{figure*}[th]
\centerline{\includegraphics[width=0.95\textwidth]{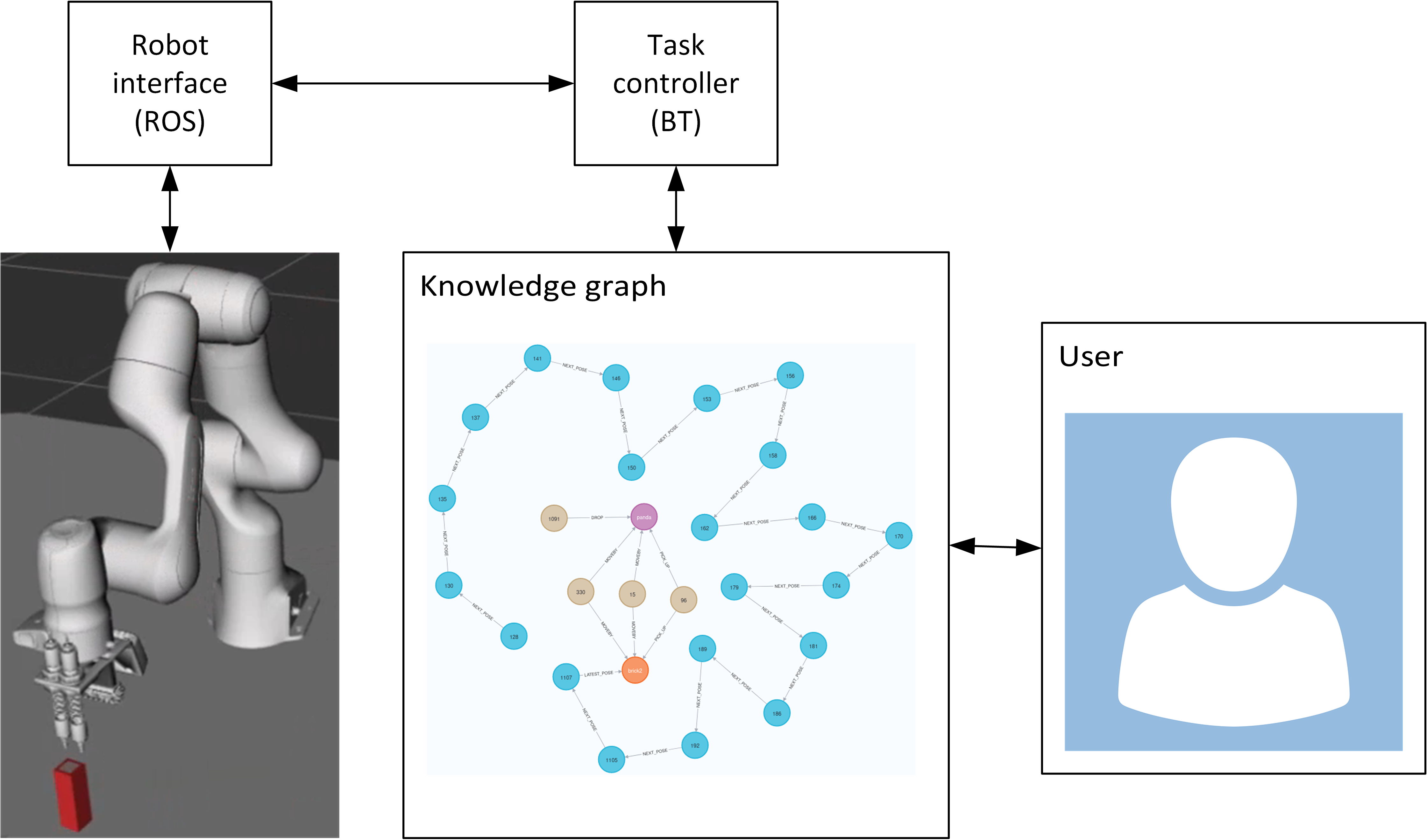}}
\caption{Overview of system. The user interacts with a robot by adding new task event nodes to a knowledge graph. These are contextualized by automatically linking them to existing agents and objects in the KG. A task controller periodically polls the KG for new task events, extracts the necessary context information to setup the skills required and coordinates the execution by the robot using ROS.}
\label{overview}
\end{figure*}

Fig.~\ref{overview} shows an overview of the proposed instructions-based programming framework. In order to instruct the robot how to perform a complicated task, the user proceeds step-by-step making use of a library of skills to accomplish simpler sub-tasks. Each step consists of three parts: skill selection, skill execution and revision. Hence, the user starts by selecting a skill from the library. For example, to manipulate the red brick on the picture, the first sub-task may be to position the suction tip of the vacuum gripper at a certain distance on top of the brick. This can be accomplished by calling a perception skill to locate an affordance on the brick, followed by a motion skill to position the suction tip of the end effector at the identified affordance point. After each skill is executed by the robot, the user can observe the outcome and decide if the result was successful. If not, the user may choose to modify the parameters and repeat the previous step. Note that this would be simple to do in a simulation environment but more complicated in the real world, where manually resetting the robot and the objects in the workspace may be needed. In the proposed framework, the state of the robot and the environment are stored in a skill-centric knowledge graph (SKG). As shown in the figure, the user interacts with the robot through the SKG, where the instructions from the user are inserted as time-stamped programming event nodes, with the necessary relationships linking the events with the robot and the manipulated objects. The task controller polls the SKG for new sub-tasks to execute. When a new event occurs, it controls the robot to execute the task using the skills from a library of skills, that is obtained from the SKG at the start of a programming session. As the task proceeds, the user can add new skills to the library by selecting useful sequences of sub-tasks. As described in Fig.~\ref{main-bt} and Fig.~\ref{base-skill-bt}, not only the task controller is implemented as a behaviour tree, but exploiting the modularity of BTs, the skills are also BTs, with low-level controllers at the leaf nodes. The interface between the robot and the task controller is via ROS~\cite{mower2024ros}.

\section{Skill-centric Knowledge Graph}
\begin{figure}[htbp]
\centerline{\includegraphics[width=0.45\textwidth]{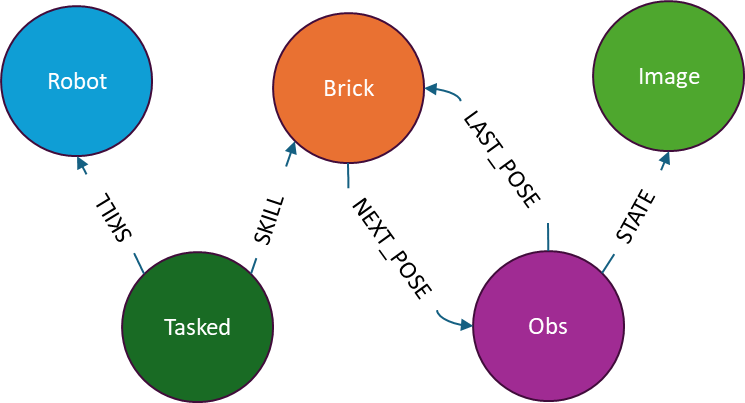}}
\caption{Base nodes and relations in Skills-centered Knowledge Graph.}
\label{tasked}
\end{figure}

Knowledge graphs are used as explicit representations of the system and contextual information that typically would exist as implicit information distributed throughout the system, such as parameters in the task controller. In a pick and place use case, information specific to the robot or to manipulable object items are stored in the corresponding graph nodes, as shown in Table~\ref{robot_att} and Table~\ref{brick_att}, respectively. In a SKG, programming actions generate Tasked events associated to an agent that is assigned to perform a task using a skill and possibly acting on some manipulable object. The relationships between the Tasked event and the agents and objects involved are captured by the edges connecting the appropriate nodes, as illustrated in Fig.~\ref{tasked} and Table~\ref{event_att}. Dynamic information from sensors generate Observed events that involve certain agents and objects captured by relationships in the SKG. As shown in Fig.~\ref{tasked}, multiple observations of the same agent or object are interconnected based on time stamps. Hence, all the contextual information associated with a Task can be obtained simply by performing path traversals in the SKG.

\begin{figure}[htbp]
\centerline{\includegraphics[width=0.45\textwidth]{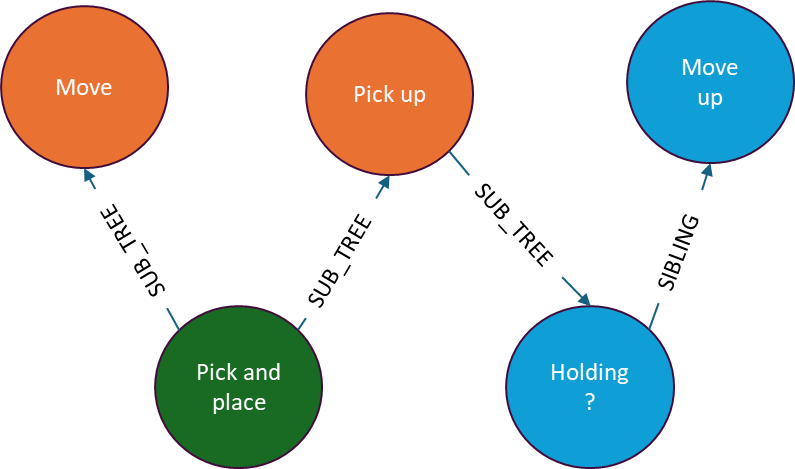}}
\caption{Snippet of a Pick and Place skills library stored in a Knowledge Graph.}
\label{pnp-library}
\end{figure}

Skills are also stored in the SKG as shown in Fig.~\ref{pnp-library}. The figure illustrates a simple pick and place library consisting of two skills, Move and Pick up. The latter is a sub tree with two branch nodes, a Holding? condition and a Move up skill. Each node contains parameters, representing the style, velocity or other variables that define the skill.

\begin{table}[htbp]
\caption{Brick attributes}
\begin{center}
\begin{tabular}{|l|l|}
\hline
\multicolumn{2}{|l|}{\textbf{Brick}} \\
\cline{1-2}
Element Id & 1 \\
Name & brick1 \\
Color & red \\
Centroid & [0, 0, 0] \\
Affordances & [[0, 0, 0.05], … [0.02, 0.02, 0]] \\
Mesh & [[0.02, 0.02, 0.05], … [0.02, 0.02, -0.05]] \\
\hline
\end{tabular}
\label{brick_att}
\end{center}
\end{table}

\begin{table}[htbp]
\caption{Robot attributes}
\begin{center}
\begin{tabular}{|l|l|}
\hline
\multicolumn{2}{|l|}{\textbf{Robot}} \\
\cline{1-2}
Element Id & 2 \\
Name & Panda \\
Tip Orientation & [0, 0, 0, 1] \\
Tip Translation & [0, 0, 0] \\
\hline
\end{tabular}
\label{robot_att}
\end{center}
\end{table}

\begin{table}[htbp]
\caption{Events attributes}
\begin{center}
\begin{tabular}{|l|l|}
\hline
\multicolumn{2}{|l|}{\textbf{Event}} \\
\cline{1-2}
Element Id & 3 \\
Name & Tasked \\
Time stamp & 2024-Jul-12-12-00-00.000 \\
Signature & Show\_last\_n\_tasks() \\
\hline
\end{tabular}
\label{event_att}
\end{center}
\end{table}

\section{Task controller}

\begin{figure}[htbp]
\centerline{\includegraphics[width=0.45\textwidth]{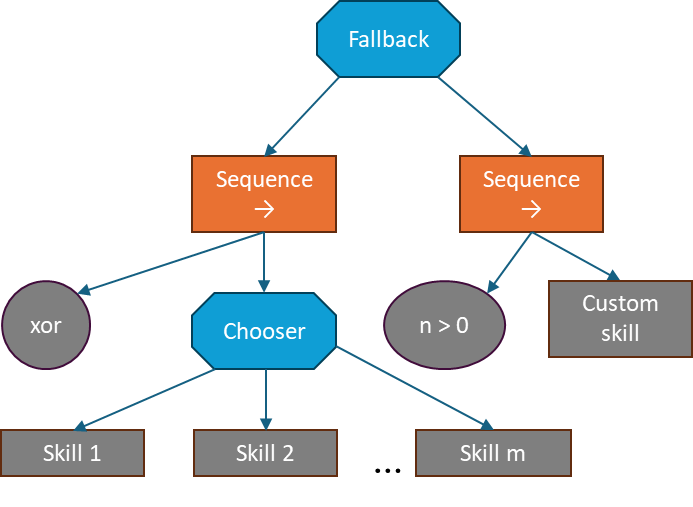}}
\caption{Behaviour tree with base skills and custom skill. The Chooser sub-tree returns Success when the xor condition is satisfied. The Custom skill sub-tree is ticked only in case the Chooser sub-tree returns Failure. }
\label{main-bt}
\end{figure}

\begin{figure}[htbp]
\centerline{\includegraphics[width=0.45\textwidth]{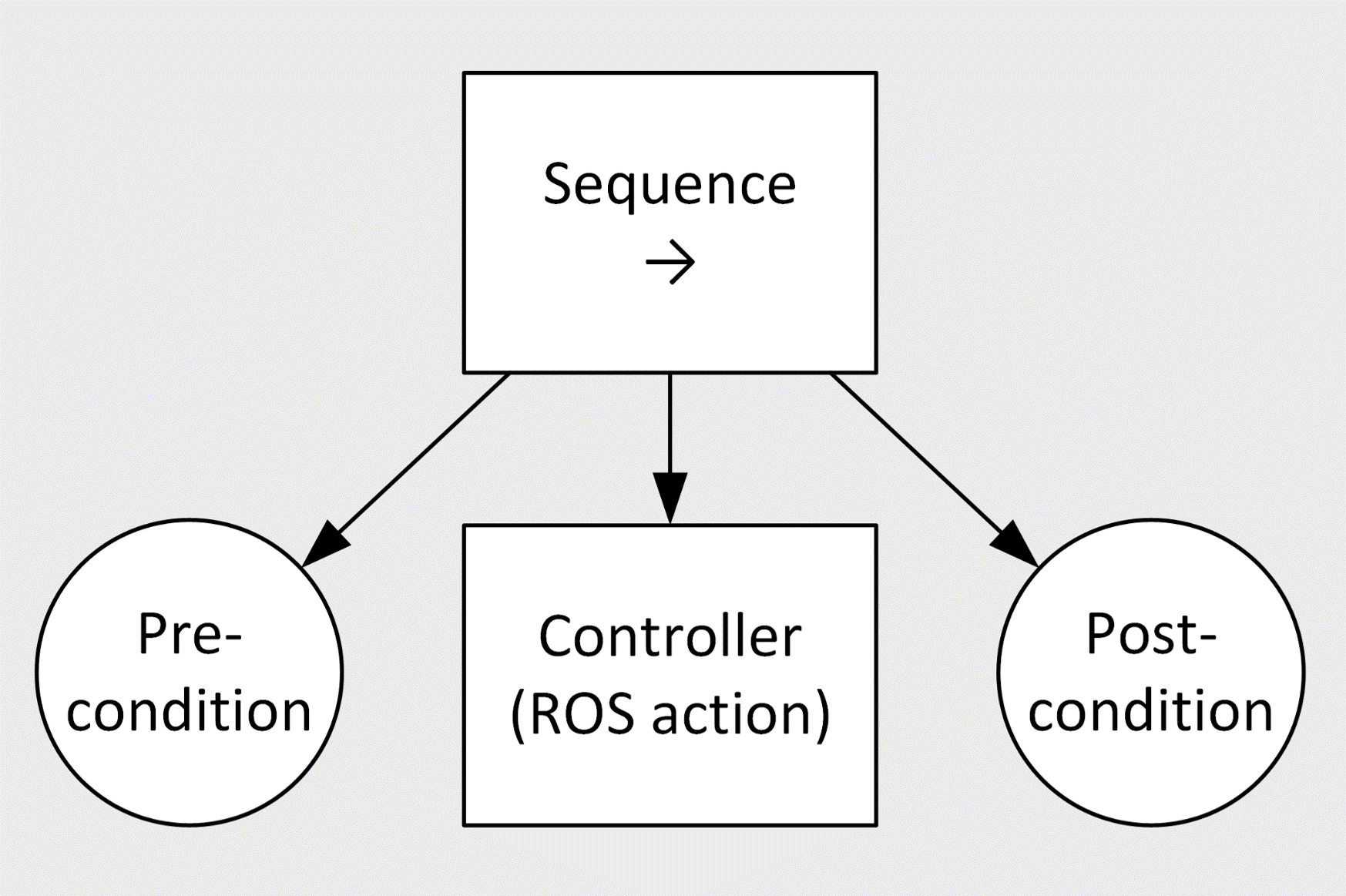}}
\caption{BT of base skill.}
\label{base-skill-bt}
\end{figure}

Fig.~\ref{main-bt} shows a block diagram of our main BT. The Fallback node at the root of the tree selects at least one of the two subtrees on every control cycle. The left subtree has the highest priority and is always executed, while the right subtree is only executed in case the left subtree returns Failure. The left subtree always returns Failure when the x-or condition is not satisfied. The structure with the x-or conditional node followed by a chooser node follows the ``Either Or" sub-tree selector implementation in py\_trees~\footnote{$https://github.com/splintered$-$reality/py\_trees$}. The x-or node ensures that the chooser is enabled only when exactly one of its children is enabled by setting the corresponding flag on the chooser structure to True, while all other flags are False. So, Skill 1 is executed when only the first flag is True, and so on. Internally, each skill is itself implemented as a BT, with at least one leaf node corresponding to a Base Skill, as shown in Fig.~\ref{base-skill-bt}.

The Base Skill returns Success when the Post-condition is satisfied; it returns Failure when any of its children returns Failure. When the Pre-condition is satisfied, the Controller initializes and attempts to run the corresponding skill by calling a ROS action server. For example, in a pick and place scenario, at least three base skills are needed as shown in Table~\ref{basic_skills}, to move the end effector from its current position to a goal, to grasp and release a payload and to locate a manipulable object, respectively. To initialize the controller, task-related parameters obtained from the SKG are used to setup the goal. Note that the same controller, setup with different parameters can perform multiple tasks, or perform a task with different styles. For example, the pose of the end effector relative to a manipulable object can be set normal to the surface to pick up the object from a table, or obliquely to scan a QR code on the surface of the object, while in both cases the same Cartesian controller is used.

\begin{table}[htbp]
\caption{Base skills}
\begin{center}
\begin{tabular}{|l|l|}
\hline
\multicolumn{2}{|l|}{\textbf{Skill}} \\
\cline{1-2}
Cartesian & Move end effector in cartesian space \\
Gripper & Open or close gripper \\
Perception & Locate affordances of objects \\
\hline
\end{tabular}
\label{basic_skills}
\end{center}
\end{table}

The base skills are normally defined by experts, as they require a deeper understanding of perception algorithms, kinodynamic constraints of the robot, human-robot interaction principles, etc. Nevertheless, having the ability to extend the library even by non-experts is fundamental to enable efficient programming by instructions. In order to do this, the BT needs to handle arbitrary sequences of parameterized actions defined by sets of instructions issued by the programmer. 
We introduce the Custom Skill node for this purpose. As shown in Fig.~\ref{main-bt}, the Custom skill subtree is selected by the Fallback node only when the Chooser subtree returns Failure. This happens when the x-or condition is not satisfied because there are no base skills either running, or scheduled to run. When a new custom skill is scheduled, the counter $n$ is initialized to the number of Base skills that will be called by the Custom skill node. This makes the condition $n>0$ True and the Custom skill is selected. Algorithm~\ref{alg:custom} describes the functionality of the custom skill node assuming the basic skills are MOVE, GRIPPER and PERCEPTION. When the node is initialized by the task controller, the $step$ counter is reset. Each time the node is ticked, the base skill corresponding to the current $step$ is initialized and enabled, such that in the next control cycle it gets selected by the task controller. After this skill finishes, the custom skill is selected again, repeating the same process until the $step$ counter reaches the number of steps in the custom task.

\begin{algorithm}
\label{alg:custom}
\small{
\SetAlgoLined
  \console{ClearChooserFlags()} \\
  \If{\console{\step} >= len(Task)}
  {
    {\bf return} True
  }
  \console{task}$\leftarrow$Task[\step] \\
  \console{data}$\leftarrow$Data[\step] \\
  \console{\step}$\leftarrow$\step $+1$ \\
  \eIf{
    \console{task} == \console{MOVE} |
    \console{task} == \console{GRIPPER}  |
    \console{task} == \console{PERCEPTION}
  }
  {
    \console{SetUpTask(task, data)}
  }
  {
    {\bf return} False
  }
  {\bf return} True
}
\caption{\console{Custom Skill}}
\end{algorithm}

\section{Results}
To test the proposed system, we created a realistic pick and place simulation in Drake~\cite{drake}. 

\subsection{Simulation environment}

A Franka-Emika Panda 7 DoF manipulator with a custom gripper is used. The robot can be controlled by specifying a target pose for the end effector. In our current implementation, motion skills are implemented as minimum jerk trajectories in Cartesian space, assuming all time-derivatives are zero at the start and goal positions~\cite{flash1985coordination}. In simulation, the translational and angular velocities required by the minimum jerk polynomials are tracked using the differential inverse kinematics controller in Drake, without collision avoidance. The custom gripper integrates a RealSense camera, a QR code scanner and a vacuum gripper. Grasping is simulated based on the distance between the tip of the vacuum gripper and the nearest point on the surface of the manipulable objects and time. Whenever the gripper is turned on, when the distance first crosses a distance threshold a timer is started. If the distance remains below the threshold after the timer has run for a time duration, a virtual joint connecting the object and the gripper is locked and the object is grasped. To release the object, after the gripper is turned off for a certain time duration, the joint is unlocked. 

The robot work space is observed using a fixed RealSense camera. It consists of a table with six manipulable bricks of different colors and sizes. Each brick has a pre-specified set of affordances. While the cameras could be used for detection and segmentation, we forego grasp modeling and extract object poses from the simulator. The afforrdance point on an object farthest from the surface of the table along the vertical is selected as the key point for grasping as well as for object-centric moves. 

\subsection{Task controller}
We implement a BT in python using py\_trees. The robot is capable of the basic move, gripper and perception skills defined in Table~\ref{basic_skills}. Based on this basic set, we added four useful skills for pick and place: 1) pickup\_brick, 2) drop\_brick, 3) move\_hand and 4) move\_by\_object. The latter moves the end effector relative to a selected reference object. The drop primitive is a combination of move and gripper. The pick up primitive is a combination of move, gripper and perception. The transitions between the states are controlled by a behaviour tree. Finally, we add a custom skill, which is a sequence of any number of base skills, possibly with condition checks. This leverages object-centric skills, which allow applying the same sequence of skills to different objects. The programmer selects between one of the three base skills or the custom skill. The BT executes the skill until it either completes the task, fails or is interrupted by a stop command.

\subsection{Skill-centric knowledge graph}
Information about the system and the state of the task are stored in a Knowledge graph implemented in Neo4j. Information about the system includes the number of blocks, their colors, mesh points and affordances. Information about the state of the task includes the pose of the robot, the pose of the blocks, the pose of the gripper, the state of the gripper, the state of the vacuum and the task. The state is captured by the internal sensors of the robot and by the static RS camera. The SKG stores the time-stamped state information. The nodes of the graph include agents, i.e., the robot, blocks, tasks and observations. The edges of the graph include the relationships between the various entities.

\subsection{Programming interface}
The user interacts with the robot by calling simple python commands. These are a few examples.

\begin{minted}[breaklines, breakafter=d]{python}

# Pick up a brick
pickup_brick('red', offset=3)
# Drop the brick
drop_brick(orientation=[0,0,0], offset=3)
# Move the end effector
move_hand(orientation=[0,0,-90], translation=[0,20,0])
# Move end effector relative to a brick
move_by_object('blue', translation=[0,0,-5])
# Save the last-n tasks as named skill
save_last_n_tasks('PileBrickOnBrick', 7)
# Do named skill from library
do_skill_from_library('TipOverBrick', {'red': 'green'})
# Show last-n tasks
show_last_n_tasks(10)

\end{minted}

The complete system, runs in a laptop with an Intel Core i7 CPU and 16 GB of memory.

\subsection{Experimental results}
\begin{figure*}[t]
\centerline{\includegraphics[width=0.95\textwidth]{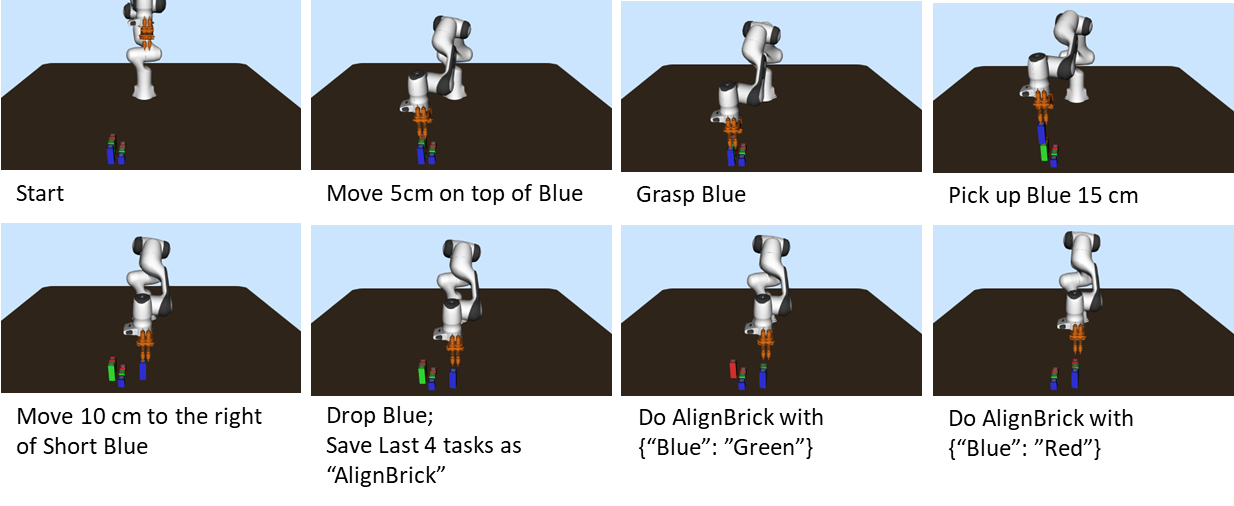}}
\caption{Experimental results.}
\label{experiment}
\end{figure*}

Fig.~\ref{experiment} shows an example of the system running in simulation. For this example, rather than calling the skills by entering text, e.g. in an interactive python notebook, we used a pre-trained language model to generate the necessary python code from natural language, similar to the labels under each frame. Each frame shows the state of the simulation, as captured by the camera after executing the skill represented by the caption under the frame. Specifically, the end goal was to move the row of three tall bricks 10 cm to the right of the short bricks. We provided instructions to show how to move one brick using object-centric motion, plus the necessary grasp and release gripper commands. We recorded the instruction in the SKG as a new skill called AlignBrick, and then applied the new skill to the other two bricks.

\section{Conclusions}

We proposed an interactive system for programming by instructions. We anticipate that using this system, non-robotics experts should be able to program a robot to perform complicated tasks in an operational domain. Furthermore, as we have shown, this can be done using natural language to simplify the programming task. We are currently working on extending the use of large language and vision models to support instruction generation, and on evaluating this system in real world use cases in semiconductor manufacturing and validation.

\section*{Acknowledgment}

Thanks to the Human Robot Collaboration team in Intel Labs for the sharing of ideas and discussions that lead to this paper.

\bibliographystyle{IEEEtran}
\bibliography{biblio}

\end{document}